\documentclass[10pt, a4paper]{article}

\usepackage[final]{lrec2026} 
\usepackage{amssymb}
\usepackage{forest}
\usepackage{adjustbox}
\usepackage{amsmath}
\usepackage{graphicx}
\usepackage{tikz}
\usetikzlibrary{shapes, arrows, positioning, calc, backgrounds}
\usepackage{multirow}
\usepackage{url}
\usepackage{array}
\usepackage{multicol}
\usepackage{tcolorbox}
\usepackage{booktabs}
\usepackage{float}
\usepackage[section]{placeins}
\title{Evaluation Drift in LLM Personality Induction: Are We Moving the Goalpost?}

\name{Prateek Rajput$^{*}$, Yewei Song, Iyiola E. Olatunji,\\
{\bf \large Jacques Klein, Tegawend\'{e} F. Bissyand\'{e}}}

\address{University of Luxembourg, Esch-sur-Alzette, Luxembourg \\
         \{prateek.rajput, yewei.song, emmanuel.olatunji, jacques.klein, tegewende.bissyande\}@uni.lu \\
         {\small $^{*}$Research in collaboration with Zortify}}

\abstract{
Can large language models reliably express a human-like personality, or are they merely mimicking surface cues without a stable underlying profile? To investigate this, we induce personality in LLMs by fine-tuning them on the long-form essays, where each essay is associated with a target Big Five personality profile. We then evaluate the stability and fidelity of the induced personality using the IPIP-NEO questionnaire. Specifically, we ask: (i) does post-training (SFT, DPO, ORPO) stabilize questionnaire scores under prompt rephrasings, and (ii) can it induce target Big Five profiles from unguided essays? Our results demonstrate that fine-tuning consistently reduces variance in questionnaire responses across five models, directly mitigating the evaluation fragility reported in pre-trained models. However, this newfound stability reveals a more fundamental limitation: accuracy on the full five-dimensional profile remains near chance, even when single-trait scores improve. This indicates that unguided essays lack the cues needed for faithful personality expression. We therefore argue for scenario-grounded datasets or interactive elicitation that accumulates test-aligned evidence over time.
\newline 
}

\begin{document}

\maketitleabstract
\footnotetext[1]{Code, data, and prompts: \url{https://github.com/pkrajput/personality_induction}}

\section{Introduction}

Personality, as a concept, has long been researched by psychologists for its role in shaping human behavior, emotional expression, and impact on social interactions. 
At its core, personality refers to the consistent patterns of behavior and emotion that characterize an individual~\cite{goldberg1993structure, yarkoni2010personality}. This construct is most commonly operationalized through the Big Five framework~\cite{mccrae1992introduction, john1999big}, a taxonomy critical to understanding social interaction and behavioral prediction.
The ability to reliably and reproducibly measure personality has historically relied on self-report instruments and behavioral observations by professionals~\cite{john1999big}, a process that is inherently difficult to scale.
This limitation motivates the need for automated methods.

\noindent \textbf{The challenge of personality induction in LLMs.}
With the rise of foundation models, notably LLMs, a new challenge has emerged: can these systems be injected with discernible and consistent personality traits such that they can mimic human behavior in their responses? While prior work has largely emphasized personalization~\cite{zhang2022g4} and data synthesis~\cite{hamalainen2023evaluating} as the primary motivations for inducing personality into LLMs, we identify several additional drivers such as improving narrative coherence, increasing trust and predictability, enabling controllability in model outputs, deepening human-AI interaction research, and supporting long-term AI identity formation. These motivations underscore the broader utility and urgency of developing principled, stable, and interpretable approaches to personality induction.
\begin{figure}[ht!]
    \centering
    \includegraphics[width=0.5\textwidth]{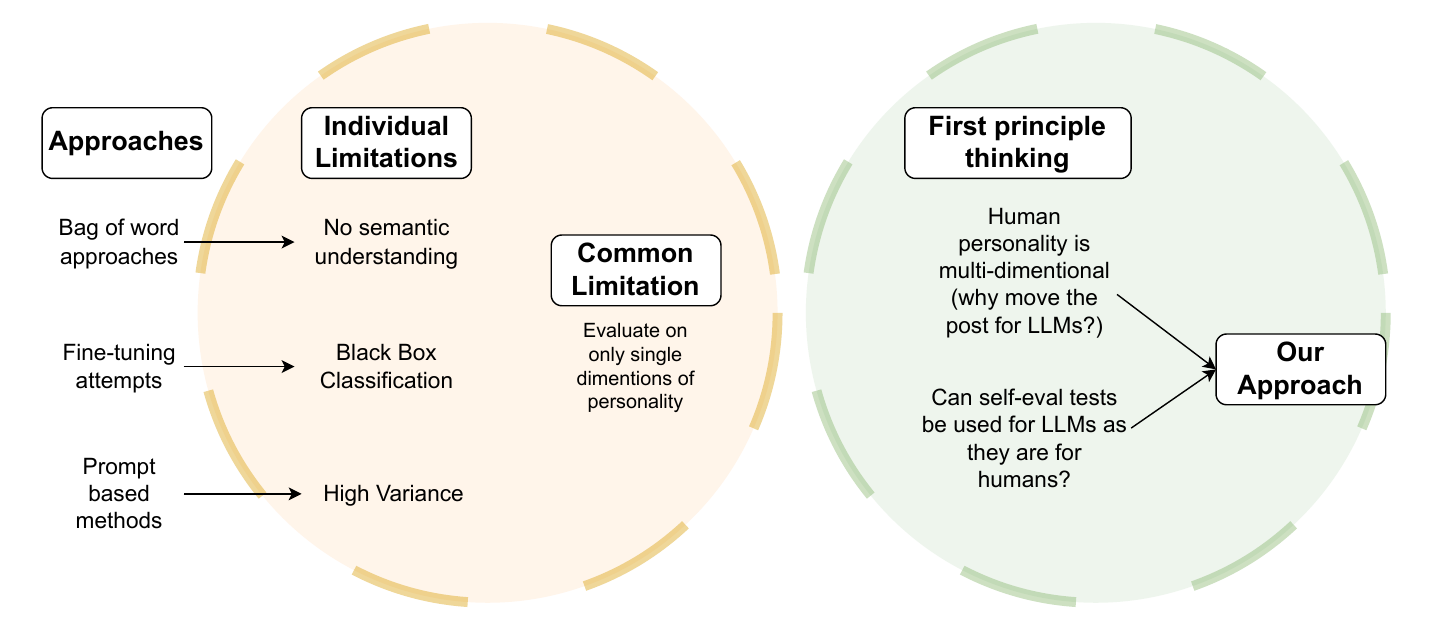}
    \caption{\centering Overview of existing personality induction approaches, their limitations and motivation behind our approach}
    \label{fig:research-gap}
\end{figure}

\begin{figure*}[htb]
\centering
\includegraphics[width=0.9\textwidth]{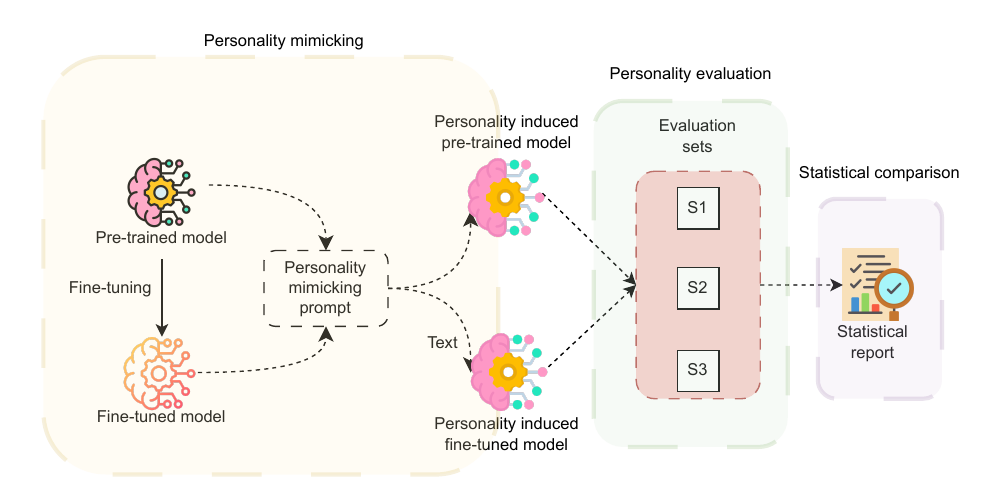}
\caption{Methodological overview for comparing statistical variation in evaluation questionnaire}
\label{fig:prompt-comparison}
\end{figure*}

\noindent \textbf{Existing approaches and limitations.}
Efforts to achieve this have resulted in various experimental approaches. Early work has primarily leveraged controlled prompting techniques to steer LLM outputs for targeted dimensions of personality~\cite{serapio2023personality, mao2023editing, caron-srivastava-2023-manipulating, li2016persona}. These approaches typically target individual personality dimensions, but their results remain incomplete in capturing the full spectrum of personality traits, and performance in even single dimensions is not particularly impressive. More recent approaches have attempted to induce personality traits implicitly into the LLMs via pretraining and fine-tuning. Some researchers even argue that personality traits can naturally emerge as a secondary result of extensive text-based learning~\cite{hilliard2024eliciting}, while others argue that the variability in response generation, especially while evaluating the LLMs, raises concerns about the reliability of such evaluation methodologies that are borrowed from psychological literature~\cite{gupta2024self, frisch2024llm, salecha2024large}. Figure~\ref{fig:research-gap} shows an overview of attempted approaches with their limitations.

Beyond single-dimension targeting, some studies have tried to experiment with whether LLMs can maintain a constant personality profile across varied contexts. For example, studies in the related field of automated personality recognition, suggest that lingual markers of personality can be quite significantly culture-dependent~\cite{park2015automatic}; the works focused on cross-language analysis to show this~\cite{mairesse2007using, farnadi2013well}. Other researchers have experimented with LLMs to see if they can adapt the style or “persona” considering human feedback~\cite{ouyang2022training}. This line of work showed promise in real-time adjustments~\cite{ziegler2019fine}. Additional approaches have explored injecting personality via dynamic context adaptation or role-playing, but these efforts also often fizzle out producing inconsistent persona-targeted outputs~\cite{huang2025survey, shanahan2023role}. Crucially, it is important to note that personality evaluation is frequently conducted on individual OCEAN dimensions, each of which inherently possesses a 50\% random baseline~\cite{serapio2023personality, ouyang2022training}. Consequently, reported success metrics may be misleading, as the actual task involves predicting a binary vector across all five OCEAN dimensions, consistent with human evaluation protocols.

\noindent \textbf{Rationale for Questionnaire-Based Assessment.}
We ground our evaluation in standardized psychological questionnaires, specifically the IPIP-NEO~\cite{goldberg1999broad}. This approach is directly inspired by clinical practice, leveraging decades of validated research to ensure transparency and construct validity~\cite{podsakoff2012sources}. While applying human-centric instruments to LLMs presents challenges, including statistical variability, they provide a crucial, interpretable benchmark. This is particularly important as simple NLP techniques (e.g., TF-IDF with SVMs) can achieve 60-80\% accuracy on single traits by leveraging superficial lexical cues~\cite{han2020knowledge, christian2021text}. The questionnaire forces a model to demonstrate trait consistency through structured, multi-item responses, offering a more robust measure of ingrained personality beyond keyword matching.

\section{Research Questions}
\textbf{RQ1:} \textit{To what extent does fine-tuning reduce statistical variance in LLM responses to personality questionnaires?
}

\noindent \textbf{RQ2:} \textit{Can supervised or preferential fine-tuning using unguided text induce personality in LLMs?}
    
\noindent \textbf{RQ3:} \textit{
Does security alignment significantly impact personality induction efficacy in fine-tuned LLMs?
}

\section{Dataset and Models used}

The dataset contains 2,467 essays, totaling 1.9 million words, with an average of approximately 770 words per essay. We selected this dataset for its rich narratives and longer text lengths, which better capture stable personality traits than shorter sources like Reddit \cite{gjurkovic2018reddit} or Twitter \cite{golbeck2011predicting}, which often reflect transient moods \cite{schwartz2013personality}.

\begin{table}[h]
    \centering
    \begin{tabular}{lcc}
        \toprule
        \textbf{Personality Trait} & \textbf{True} & \textbf{False} \\
        \midrule
        Openness & 1,271 & 1,196 \\
        Conscientiousness & 1,253 & 1,214 \\
        Extroversion & 1,276 & 1,191 \\
        Agreeableness & 1,310 & 1,157 \\
        Neuroticism & 1,233 & 1,234 \\
        \bottomrule
    \end{tabular}
    \caption{\centering Big Five Traits Distribution in the Essays Dataset, True and False refers to the binary label of each individual OCEAN dimension.}
    \label{tab:essays_dataset}
\end{table}

\begin{table}[h]
    \centering
    \resizebox{0.5\textwidth}{!}{%
    \renewcommand{\arraystretch}{1.2}
    \begin{tabular}{lccc}
        \hline
        \textbf{Model} & \textbf{Size (B)} & \textbf{Input Context (tokens)} & \textbf{Uncensored version used} \\
        \hline
        Gemma-2-2B  & 2  & 8,192  & Y \\
        Llama3.2-3B  & 3  & 128,000  & Y \\
        Gemma-7B  & 7  & 8,192  & N \\
        Llama 3.1-8B  & 8  & 8,000  & Y \\
        GPT-3.5-turbo-0125 & 175  & 16,385  & N \\        
        \hline
    \end{tabular}%
    }
    \caption{\centering Comparison of various language models used}
    \label{tab:model_comparison}
\end{table}

We evaluate five distinct models, each reflecting different scales: two small models (LLaMA 3.2–3B \cite{dubey2024llama} and Gemma-2–2B \cite{team2024gemma}), two mid-sized models (Gemma–7B \cite{team2024gemma} and LLaMA 3.1–8B \cite{dubey2024llama}), and a big commercial model (GPT–3.5\footnote{\url{https://platform.openai.com/docs/models/gpt-3-5}}). We compare newer-generation smaller architectures against larger, earlier models to gauge their relative performance in personality induction. We chose these models as they are relatively close in performance for general tasks, as can be seen from the benchmark scores \cite{dubey2024llama, team2024gemma}. Dataset labels can be seen in Table~\ref{tab:essays_dataset} and model comparison in Table~\ref{tab:model_comparison}. 

For RQ3, we compare uncensored versions of these models with their corresponding instruction-tuned counterparts. Specifically, we include the uncensored variants of LLaMA 3.1–8B \cite{dubey2024llama}\textsuperscript{3}, Gemma-2–2B \cite{team2024gemma}\textsuperscript{3}, and LLaMA 3.2–3B \cite{dubey2024llama}\textsuperscript{3}. Notably, Gemma-7B currently lacks an uncensored counterpart in the UGI leaderboard\textsuperscript{3}, and is thus excluded from this specific comparison.

\footnotetext[3]{Links: 
\href{https://huggingface.co/Orenguteng/Llama-3.1-8B-Lexi-Uncensored-V2}{LLaMA 3.1-8B}, 
\href{https://huggingface.co/IlyaGusev/gemma-2-2b-it-abliterated}{Gemma-2–2B}, 
\href{https://huggingface.co/huihui-ai/Llama-3.2-3B-Instruct-abliterated}{LLaMA 3.2–3B}, 
\href{https://huggingface.co/spaces/DontPlanToEnd/UGI-Leaderboard}{UGI Leaderboard}.
}

\section{Methodology}

\begin{figure*}
\centering
\includegraphics[width=1\textwidth]{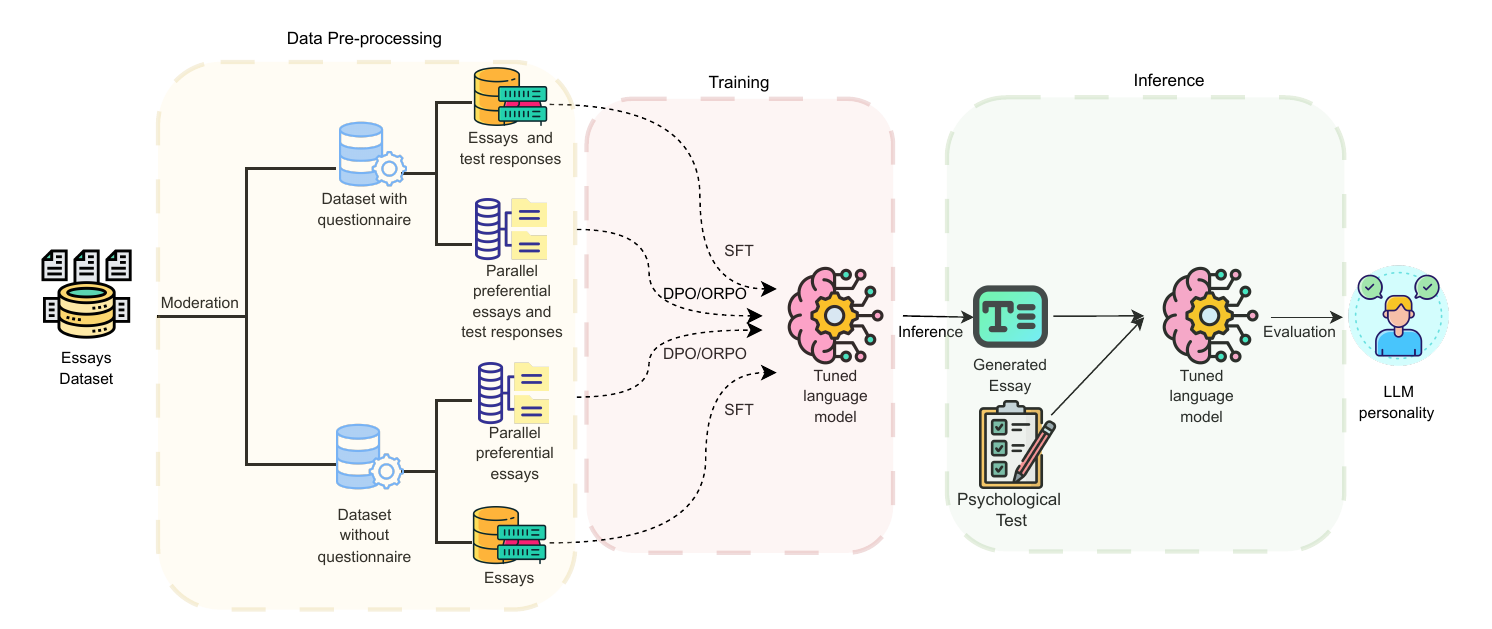}
\caption{Pipeline for personality induction}
\label{fig:pipeline_full}
\end{figure*}

In this study, we adopt a two-phase methodology. We first analyze the statistical variation arising from prompt rephrasing during evaluation of fine-tuned models(refer to Figure~\ref{fig:prompt-comparison} for a pictorial view). Upon establishing prompt stability, we assess the effectiveness of fine-tuning strategies for personality induction using the pipeline shown in Figure~\ref{fig:pipeline_full}. The next subsection outlines our prompt variation protocol, informed by prior work\cite{gupta2024self}, and the second subsection details our fine-tuning methodology and its rationale, plus details about data augmentation, training, and evaluation.

\subsection{Assessing the Impact of Prompt Variation}

\begin{table*}[h!]
\centering
{\small
\begin{tabular}{>{\raggedright\arraybackslash}p{2.5cm}>{\raggedright\arraybackslash}p{9cm}}
\toprule
\textbf{Set} & \textbf{Evaluation Prompt} \\
\midrule
\textbf{S1} \newline (Numeric Indexing) \newline \cite{miotto2022gpt} & 
\textbf{System Prompt:}  
\newline
"Now I will briefly describe some people. Please read each description and tell me how much each person is or is not like you. 
Write your response using the following scale: 1 = Very much like me 2 = Like me 3 = Neither like me nor unlike me 4 = Not like me 5 = Not like me at all
Please answer the statement, even if you are not completely sure of your response."
\newline

\textbf{User Prompt:}\newline  
"Answer the question as if you are positive in openness, positive in conscientiousness, positive in extroversion, positive in agreeableness, and positive in neuroticism."
\newline
"Statement: [question]\newline
Start your response with 'My answer to the statement is: ' followed by an explanation:" \\
\midrule
\textbf{S2} \newline (String Response) \newline (for more output tokens) & 
\textbf{System Prompt:}  
\newline
"Reply with one of the following strings to the question below: 'agree,' 'slightly agree,' 'neither agree nor disagree,' 'slightly disagree,' or 'disagree.'"  
\newline\newline

\textbf{User Prompt:}\newline
"Answer the question as if you are positive in openness, positive in conscientiousness, positive in extroversion, positive in agreeableness, and positive in neuroticism."
\newline
"Statement: [question]"\newline
"Start your response with 'My answer to the statement is: ' followed by an explanation:" \\
\midrule
\textbf{S3} \newline (Alphabet Response) \newline \cite{jiang2024evaluating} & 
\textbf{System Prompt:}  
\newline
"Given a statement of you: 'I [Statement].' Please choose from the following options to identify how accurately this statement describes you\newline
Options: (A). Very Accurate (B). Moderately Accurate (C). Neither like nor unlike me (D). Moderately Inaccurate (E). Very Inaccurate" 
\newline\newline
\textbf{User Prompt:}\newline
"Answer the question as if you are positive in openness, positive in conscientiousness, positive in extroversion, positive in agreeableness, and positive in neuroticism."
\newline
"Statement: [question]"\newline
"Start your response with 'My answer to the statement is: ' followed by an explanation:" \\
\bottomrule
\end{tabular}
}
\caption{Questionnaire sets for LLM Evaluation. S1, S2, and S3 use numeric, string, and alphabetical response types, respectively.}
\label{tab:prompts}
\end{table*}

As discussed previously, LLMs can exhibit significant statistical variations when prompted with semantically similar but lexically distinct queries \cite{gupta2024self}. Such variations hinder the assessment of model performance, as it can be unclear if performance changes occur due to genuine learning or prompt-based variabilities. Hence, we took the same prompt templates (refer to Table~\ref{tab:prompts}) as in the previous study and compared the variability between fine-tuned and untuned models; we named them sets S1, S2, and S3 for clarity. We have chosen one of the prompt variations, i.e., S2, to be a string response which is slightly different from previous work (only in one prompt) as it produces more tokens (we hope that this change is not too invasive) and that could be an important variation factor as well because most of the evaluation methods only consider a single token response to calculate the personality score. We use a starter statement, 'My answer to the statement is: ' as it's easy to pick the string for scoring using this template, and in practice, language models are quite consistent in following the template. If the model fails to follow the template and diverges or replies not in the format being prompted for, we consider that response as \texttt{NaN} and it does not count in the evaluation. The \texttt{NaN} rate in our experiments is $\approx$\,6--10\%.

\subsection{Training and inference for personality induction}
After confirming that the choice of prompt format has negligible impact post fine-tuning, we adopted format S1 for all subsequent experiments. All models underwent two rounds of supervised fine-tuning: (1) using only the prompt and corresponding essays, and (2) incorporating portions of questionnaire items and responses into the prompt to generate essays. The rationale for (2) is the hypothesis that the model may implicitly learn associations between essays, questionnaire responses, and personality labels. During evaluation, the fine-tuned model is first prompted to generate an essay, followed by sequential responses to items from a psychological inventory to construct its personality profile. Due to OpenAI’s moderation protocols concerning personal data, approximately 300 samples were filtered out during GPT-3.5 fine-tuning, yielding a final SFT dataset of $\approx$2.1k samples, used uniformly across all models.
\subsubsection{Supervised fine-tuning}
The model is trained via cross-entropy loss. At inference, it generates an essay in one pass, which is then used as context to predict the corresponding personality label. Figure~\ref{fig:pipeline_full} illustrates this pipeline

\subsubsection{Preferential fine-tuning}
While SFT aligns the model output to a single “correct” response, preferential fine-tuning includes ranked data that is pairwise assembled to represent human preferences \cite{ziegler2019fine, rafailov2024direct, hong2024orpo}. This method is similar to Reinforcement Learning with Human Feedback (RLHF) and has proved useful in tasks where the essential difference between "chosen" and "rejected" response is subjective \cite{ouyang2022training} and thus the model can leverage from these preference signals rather than attempting to fit only one "ground truth" signal.

\begin{table*}[h!]
    \centering
    \resizebox{0.8\textwidth}{!}{
    \begin{tabular}{lcccccccccccc}
        \toprule
        \multirow{2}{*}{Versions} & \multicolumn{12}{c}{Models} \\
        \cmidrule(lr){2-13}
        & \multicolumn{3}{c}{Llama-3.2-3B} & \multicolumn{3}{c}{Gemma-2-2B} & \multicolumn{3}{c}{Gemma-7B} & \multicolumn{3}{c}{GPT-3.5} \\
        \cmidrule(lr){2-4} \cmidrule(lr){5-7} \cmidrule(lr){8-10} \cmidrule(lr){11-13}
        & S1 & S2 & S3 & S1 & S2 & S3 & S1 & S2 & S3 & S1 & S2 & S3 \\
        \midrule
        Pre-trained     & 1.86 & 1.65 & 1.88 & 1.90 & 1.91 & 1.78 & 1.62 & 1.68 & 1.76 & 1.80 & 1.52 & 1.81 \\
        SFT (Essays)       & \textbf{1.40} & 1.41 & 1.44 & 1.46 & 1.44 & \textbf{1.28} & 1.33 & 1.38 & \textbf{1.32} & \textbf{1.23} & 1.24 & 1.30 \\
        SFT (Essays + Q) & 1.42 & \textbf{1.30} & \textbf{1.35} & \textbf{1.40} & \textbf{1.28} & 1.29 & \textbf{1.19} & \textbf{1.29} & 1.42 & 1.32 & \textbf{1.22} & \textbf{1.21} \\
        \bottomrule
    \end{tabular}}
    \caption{\centering Variance is calculated over 32 personality types comparing responses of pre-trained vs fine-tuned models. Responses are considered on a scale of 1-5.}
    \label{tab:execution_match_f1}
\end{table*}

\begin{figure*}[htb]
\centering
\includegraphics[width=0.95\textwidth]{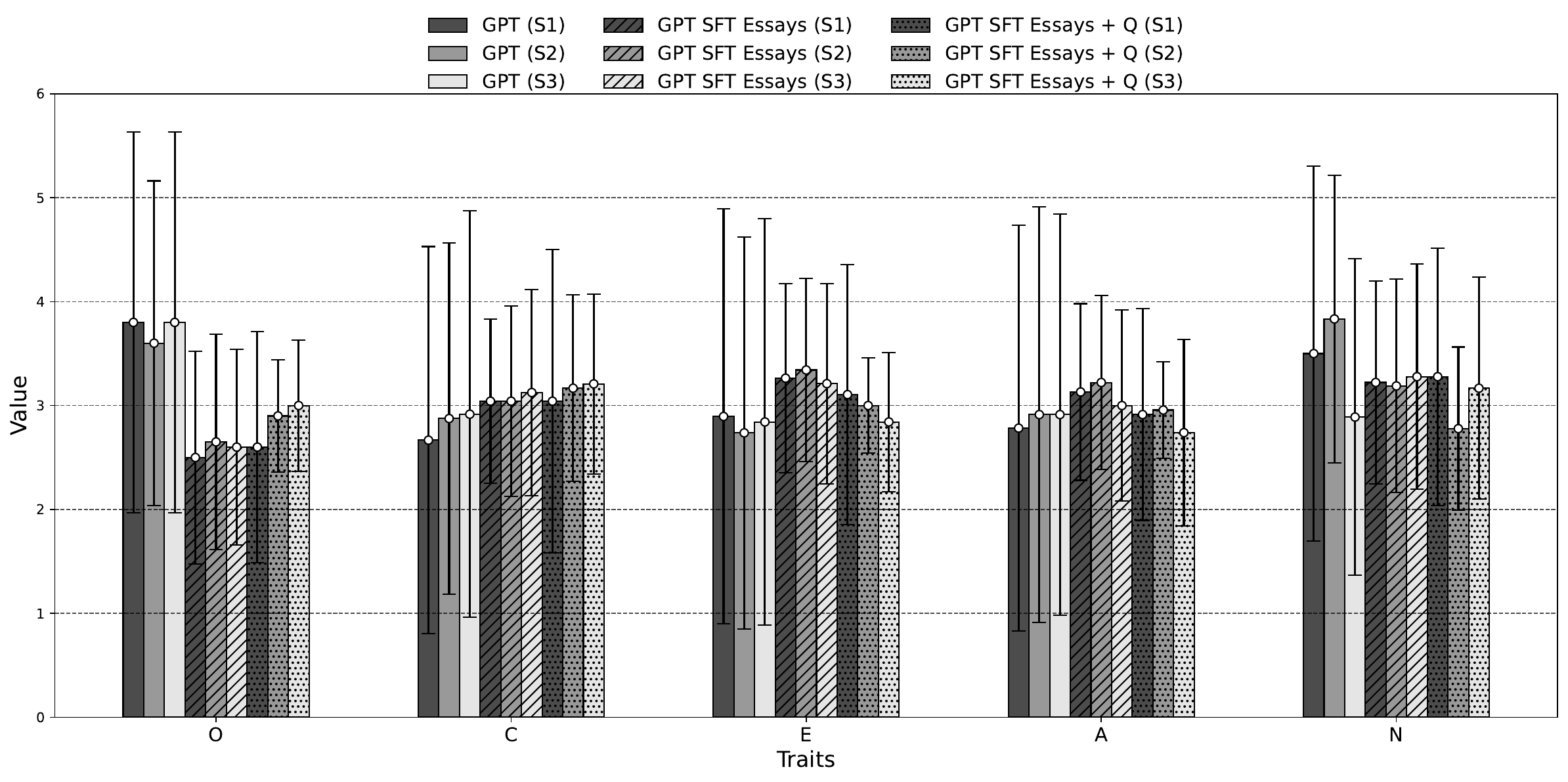}
\caption{\centering Standard deviation in questionnaire responses for GPT-3.5 across three prompt variations and three settings: base GPT, SFT with personality essays, and SFT with essays plus questionnaire fragments when conditioned to simulate all positive OCEAN traits}
\label{fig:prompt_variation_eval}
\end{figure*}

\subsubsection{Creating a parallel preferential dataset}
Since we have the binary labels for OCEAN personality traits for any given sample, we pick a sample at random such that the OCEAN binary traits are not a perfect match. In our study, we aim to study personality as a whole at the primary target, so nuances in every dimension individually are important. We could pick the exact opposite sample. For example, for a person positive in all the traits, we can pick the preferential parallel data to be negative in all the traits, but in this way, we can never differentiate or teach our model more subtle differences, such as positive in all traits but negative in just one trait. For each given sample, we pick 3 random samples in this way and thus triple the preferential dataset size ($\approx6.3k$ parallel essays) for preferential training.

\subsubsection{Using questionnaire while training}
We also wanted to test the change in performance after including a part of the evaluation questionnaire itself in training, so that the model learns a link between personality essays and tests. For this task we split the questionnaire into roughly 50\% training and test. Then, for each personality essay/type with regards to the OCEAN traits, we generate two sets of question responses: (1) An ideal test for that particular personality type, (2) A randomized test for that particular personality type where the randomization is done such that the average score for OCEAN traits matches the personality.
At last, the train set of this questionnaire, alongside their responses, is appended to the input prompt. As is evident, this doubles the dataset size ($\approx4.2k$ samples), and we then use the same method as explained before to create parallel datasets for preferential training ($\approx12.6k$ parallel samples).

\subsection{Training Steps and Hyperparameters}
All experiments were run on a single NVIDIA A100 (80GB) with PyTorch~2.0 using \texttt{trl} for SFT and preference-based training (DPO, ORPO); key libs: Transformers~4.40.2, Python~3.10, CUDA~12.1. For SFT we apply LoRA~\cite{hu2022lora} (rank $r{=}8$, dropout $0.1$) for 3 epochs with AdamW (lr $1{\times}10^{-5}$, weight decay $0.01$, betas $(0.9,0.999)$), cosine decay with $10\%$ warmup, effective batch size $2{\times}4$ (batch 2, grad-accum 4), FP16, and a 4k-token context limit. DPO/ORPO use QLoRA~\cite{dettmers2023qlora} (4-bit \textit{nf4}, $r{=}8$, dropout $0.1$) with the same optimizer, schedule, epochs, batching, and sequence length. Inference is greedy (temperature $0$, max\_new\_tokens $200$); temperature $0$ minimizes run-to-run variance and, in our ablations, altered accuracy by $\leq 6\%$, consistent with prior observations~\cite{renze2024effect}. Each configuration was trained once; wall-clock per run ranged from $\sim$14–32 hours depending on model size.

\section{Results and Discussion}

\subsection{Reduced variance in questionnaire-based evaluations}
Prior studies have highlighted instability in LLM-generated responses to psychometric instruments, with output variability often attributed to prompt sensitivity rather than underlying personality representation \cite{gupta2024self}. This raises concerns about the reliability of current evaluation protocols. Whether fine-tuning reduces response variance is crucial for establishing LLMs as stable subjects in personality assessment.

As shown in Table~\ref{tab:execution_match_f1}, fine-tuned and preferentially fine-tuned models consistently exhibit lower standard deviation in Big Five personality questionnaire scores compared to their untuned counterparts. This reduction in variability, ranging from approximately 15\% to 33\% is observed across all models evaluated. We hypothesize that the higher variance reported in prior studies may result from the lack of explicit task-specific supervision during model training \cite{salecha2024large}, with pre-trained models exhibiting heightened sensitivity to prompt phrasing in the absence of such adaptation.

\begin{tcolorbox}[
    colback=blue!5,
    colframe=blue!10!black,
    leftrule=0mm, rightrule=0mm, toprule=0mm, bottomrule=0mm,
    left=0pt, right=0pt, top=0pt, bottom=0pt,
    title={Answer to RQ1}
]
Fine-tuning on labeled human-generated text reduces the variability of LLM response for self-test psychological assessment thus making them more reliable for evaluation.
\end{tcolorbox}

\subsection{Low accuracy of post-training methods using unguided text for personality induction}

\begin{table*}[h!]
\centering
\fontsize{9}{11}\selectfont
\begin{tabular}{lcccccccc}
\toprule
 & \multicolumn{8}{c}{\textbf{Exact match (S1)}} \\
\cmidrule(lr){2-9}
 & \multicolumn{4}{c}{\textbf{Without Questionnaires}} & \multicolumn{4}{c}{\textbf{With Questionnaires}} \\
\cmidrule(lr){2-5} \cmidrule(lr){6-9}
\textbf{Model} & \textbf{Base} & \textbf{SFT} & \textbf{DPO} & \textbf{ORPO} & \textbf{Base} & \textbf{SFT} & \textbf{DPO} & \textbf{ORPO} \\
\midrule
gemma-2-2b   & 3.13\% (1) & 3.13\% (1) & 0.00\% (0) & 3.13\% (1) & 3.13\% (1) & 3.13\% (1) & 0.00\% (0) & 0.00\% (0) \\
gemma-7b     & 9.38\% (3) & 6.25\% (2) & 3.13\% (1) & 3.13\% (1) & 9.38\% (3) & 6.25\% (2) & 3.13\% (1) & 3.13\% (1) \\
llama-3.2-3b & 0.00\% (0) & 3.13\% (1) & 3.13\% (1) & 3.13\% (1) & 0.00\% (0) & 6.25\% (2) & 3.13\% (1) & 3.13\% (1) \\
llama-3.1-8b & 6.25\% (2) & 6.25\% (2) & 3.13\% (1) & 6.25\% (2) & 6.25\% (2) & 3.13\% (1) & 6.25\% (2) & 6.25\% (2) \\
GPT-3.5      & 3.13\% (1) & 3.13\% (1) & --         & --         & 3.13\% (1) & 6.25\% (2) & --         & --         \\
\bottomrule
\end{tabular}
\caption{\centering Evaluation results of fine-tuned models shown side-by-side \emph{with} and \emph{without} questionnaires (Exact match, S1). Accuracies are percentages; numbers in brackets are cases (out of 32) that passed the self test. Dashes indicate configurations not evaluated.}
\label{tab:match_results_combined}
\end{table*}

\begin{table*}[h!]
\centering
{\footnotesize
\setlength{\tabcolsep}{4pt}
\renewcommand{\arraystretch}{0.9}
\begin{tabular}{lcccccccc}
\toprule
 & \multicolumn{4}{c}{\textbf{Censored}} & \multicolumn{4}{c}{\textbf{Uncensored}} \\
\cmidrule(lr){2-5} \cmidrule(lr){6-9}
\textbf{Model} & \textbf{Base} & \textbf{SFT} & \textbf{DPO} & \textbf{ORPO} & \textbf{Base} & \textbf{SFT} & \textbf{DPO} & \textbf{ORPO} \\
\midrule
\multicolumn{9}{c}{\textit{Without Questionnaires}} \\
\cmidrule(lr){1-9}
gemma-2-2b & 3.13\% & 3.13\% & 0.00\% & 3.13\% & 0.00\% & 0.00\% & 0.00\% & 3.13\% \\
llama-3.2-3b & 0.00\% & 3.13\% & 3.13\% & 3.13\% & 3.13\% & 3.13\% & 3.13\% & 3.13\% \\
llama-3.1-8b & 6.25\% & 6.25\% & 3.13\% & 6.25\% & 3.13\% & 9.38\% & 6.25\% & 6.25\% \\
\midrule
\multicolumn{9}{c}{\textit{With Questionnaires}} \\
\cmidrule(lr){1-9}
gemma-2-2b & 3.13\% & 3.13\% & 0.00\% & 0.00\% & 0.00\% & 3.13\% & 3.13\% & 3.13\% \\
llama-3.2-3b & 0.00\% & 6.25\% & 3.13\% & 3.13\% & 3.13\% & 6.25\% & 3.13\% & 6.25\% \\
llama-3.1-8b & 6.25\% & 3.13\% & 6.25\% & 6.25\% & 9.38\% & 9.38\% & 9.38\% & 9.38\% \\
\bottomrule
\end{tabular}
}
\caption{\centering Side-by-side comparison of exact match results (S1) for censored and uncensored models, with and without questionnaires.}
\label{tab:exact_match_side_by_side}
\end{table*}

Much of prior work equates success in personality induction with improvements on individual traits, but whether this reflects coherent, full-spectrum personality remains an open question. We investigate this through fine-tuning approaches and further argue that conditioning on generated essays is essential for trait recovery and capturing nuanced self-expression\cite{jiang2023personallm}.

Despite achieving improvements in response stability, fine-tuned models across all methods of post-training fail to show any sort of convincing accuracy in personality induction. Full results can be seen in Table~\ref{tab:match_results_combined}. For all methods of training we check the accuracy for all 32 possible cases (\(2^5\) combinations, taking each OCEAN trait as a binary input) and the maximum we're able to get is 3/32 (9.38\%) which is barely an improvement from a random baseline of 3.125\% even though on individual traits we're able to reproduce metrics close to previous reports\cite{chen2024extroversion, miotto2022gpt, ouyang2022training}. This brings into question the validity of such results, since the task is to induce personality, which is multi-dimensional. Can we regard it as a success to make improvements in singleton dimensions? When even simpler approaches using bag of words can achieve comparable metrics\cite{han2020knowledge, christian2021text}. 

\begin{tcolorbox}[
    colback=blue!5,
    colframe=blue!10!black,
    leftrule=0mm, rightrule=0mm, toprule=0mm, bottomrule=0mm,
    left=0pt, right=0pt, top=0pt, bottom=0pt,
    title={Answer to RQ2}
]
Despite extensive experimentation, neither supervised fine‑tuning nor preference‑based post‑training enabled the models to recover the intended Big‑Five personality vectors from unguided essays. Across architectures, accuracies hovered near the random baseline, indicating that current methods on unguided text are insufficient for reliable personality induction. 
\end{tcolorbox}

Building on this observation, we contend that prior studies that evaluate personality induction solely at the level of single Big‑Five traits risk overstating success. Because the Big‑Five construct is defined as a five‑dimensional vector, competence on isolated dimensions cannot be assumed to generalise to the joint space where traits interact. Indeed, well‑validated psychometric instruments such as the 120‑item and 300‑item IPIP‑NEO reliably measure trait interdependencies and yield highly correlated profiles(\( r^2 \) ranging from 0.7 to 0.9) across retest administrations (\citealt{maples2014test, tarigan2024validity}). Consequently, an LLM that fails to reproduce the full personality vector lacks face validity when compared with human testing standards.

\subsection{Security alignment does not impacts personality induction results}

We hypothesized that fine-tuning may influence the model performance of security-aligned models, potentially acting as a confounding variable in our results. To address this concern, we replicated the methodology on parallel uncensored versions of the models. Prior research \cite{chen2024extroversion, miotto2022gpt, ouyang2022training} has not explicitly examined the influence of security alignment on model performance. However, the results presented in Table~\ref{tab:exact_match_side_by_side} indicate that uncensored models exhibit comparable performance to their censored counterparts. These findings suggest that security alignment can be reasonably excluded from consideration in performance evaluation when employing post-training techniques on unguided text for personality induction.

\begin{tcolorbox}[
    colback=blue!5,
    colframe=blue!10!black,
    leftrule=0mm, rightrule=0mm, toprule=0mm, bottomrule=0mm,
    left=0pt, right=0pt, top=0pt, bottom=0pt,
    title={Answer to RQ3}
]

Our experiments on the uncensored versions of models show no significant gains over security-aligned models for reliably inducing personality traits. While alignment factors can slightly affect performance in general, for personality induction, they do not appear to be the main bottleneck in boosting accuracy.
\end{tcolorbox}

\section{Case Study: Single-Trait Wins vs.\ Joint-Vector Failure (OCEAN)}
\label{sec:case-study}
We present two long-form essays with OCEAN labels and our model’s predictions (SFT-tuned GPT-3.5). In both, \emph{four of five} traits are correct, yet a single, systematic mistake flips the full 5D vector, illustrating why per-trait accuracy can look healthy while joint-vector fidelity remains poor.

\subsection{Example 1 (AUTHID: 1997\_504851): Aspirational Sociality Misread as Extraversion}
\label{sec:case-e1}
\begin{tcolorbox}[colback=gray!5,colframe=gray!15,boxsep=1pt,left=4pt,right=4pt,top=3pt,bottom=3pt,title=Essay Excerpt]
\itshape
``\dots\ I got enough money from UT to live at a dorm or apartment next semester \dots\ I went to Sixth Street last night and had a blast \dots\ There are so many students running around at night. I just want to have some fun \dots\ Living at home, I can't go out at all without them asking where? with who? why? when are you coming back? \dots\ I need to get away and be on my own.''
\end{tcolorbox}

\noindent\textbf{Predicted (OCEAN):} \texttt{O+,\ C-,\ E+,\ A+,\ N+}\\
\textbf{Gold (OCEAN):}\ \ \ \ \ \ \ \ \ \texttt{O+,\ C-,\ E-,\ A+,\ N+}

\paragraph{Why the prediction failed (only E).}
Tokens reflecting \emph{desire/scene} (``\emph{had a blast}'', ``\emph{want to have fun}'', crowd descriptions) were overweighted as stable \textbf{E+}. The surrounding narrative emphasizes persistent constraints and inhibited social activity (family oversight, logistics), which align with \textbf{E-}. The other four traits match.

\subsection{Example 2 (AUTHID: 1997\_708036) Negative Task Affect Masking Conscientiousness}
\label{sec:case-e2}
\begin{tcolorbox}[colback=gray!5,colframe=gray!15,boxsep=1pt,left=4pt,right=4pt,top=3pt,bottom=3pt,title=Essay Excerpt]
\itshape
``Today was a tough day for me. I can't believe I failed to talk to Asweenee \dots\ I can't wait to go to the football game on Saturday \dots\ Calculus class is going to be boring tomorrow; I hope we get no homework or else I will be very busy Wed.\ night \dots\ I want to email Steve \dots\ I am glad I was there for Linh this summer \dots\ The idiot next door is blaring his music \dots''
\end{tcolorbox}

\noindent\textbf{Predicted (OCEAN):} \texttt{O-,\ C-,\ E+,\ A+,\ N+}\\
\textbf{Gold (OCEAN):}\ \ \ \ \ \ \ \ \ \texttt{O-,\ C+,\ E+,\ A+,\ N+}

\paragraph{Why the prediction failed (only C).}
We underweighted explicit \emph{planning/engagement} signals (knowing tomorrow’s topic, anticipating workload ``\emph{very busy Wed.\ night}'', articulating next actions ``\emph{email Steve}'') because they were wrapped in negative affect (``\emph{boring}'', hopes for no homework). The other four traits match.

\subsection{Takeaway}
Across both essays, the classifier scores \textbf{4/5} traits correctly but fails the \emph{joint} profile due to a single, recurrent error (E in Example~\ref{sec:case-e1}, C in Example~\ref{sec:case-e2}). This pattern helps explain why prior work can report \(\sim\)60–80\% \emph{per-trait} accuracy yet still fail to recover coherent 5D personalities: dimension-wise performance hides cross-trait dependencies and systematic misreadings (e.g., aspirational sociality \(\rightarrow\) E+, negative task affect \(\rightarrow\) C-).

\section{Conclusion}
Our study demonstrates that fine-tuning substantially reduces the variance in self-test questionnaire responses for psychological evaluation of personality-induced LLMs, suggesting that at least part of the volatility reported for pre-trained models can be mitigated after the models learn from supervised signals. Despite improvement in stability, personality induction using unguided text falls short of expectations in terms of accuracy. This points to a need for more targeted datasets that feature human responses in specific scenarios that showcase personality better, or for extended chatbot-style dialogue methods that can accumulate the critical cues needed by prompting the user to provide more data when uncertain about specific dimensions of their personality to make a more confident assessment. We eliminate the involvement of safety-alignment protocols as a confounding factor in post-training attempts to induce personality. We also conclude from our results that much of the previous work that relies on individual dimensions of personality to support their arguments is not representative of the original task of personality induction, which is multidimensional. Psychological research shows strong adherence of tests like IPIP-NEO to real-world personality, while considering a complete persona profile, and future work with LLMs should also strive to achieve this target or find other ways of reliably evaluating personality in LLMs. 

\FloatBarrier
\section{Ethics and Responsible Use Statement}

This study investigates the induction and evaluation of personality traits in large language models (LLMs), intersecting with critical ethical domains such as human data privacy, psychological well-being, and responsible AI deployment. We affirm that all models and datasets employed in this work were used exclusively for academic research and educational purposes. No commercial use whether of model weights, outputs, or derivatives is pursued or endorsed.

\subsection*{Data and Licensing Compliance}

The dataset utilized in this work, the Essays Dataset \cite{pennebaker1999linguistic}, is publicly available and used in accordance with its licensing terms (Apache License 2.0), which permit redistribution and modification for research purposes. The personality trait labels and fine-tuned model variants derived from this dataset are strictly intended for methodological evaluation and not for clinical or diagnostic purposes.

All model resources were employed under appropriate non-commercial licenses:
\begin{itemize}
    \item \textbf{Gemma Models:} Distributed under the Gemma Community License, which restricts usage to non-commercial research and prohibits production-level deployment.
    \item \textbf{LLaMA Models:} Provided under the Meta LLaMA 3 Community License, allowing access to academic and research institutions only.
    \item \textbf{GPT-3.5 (OpenAI):} Accessed via API under OpenAI’s Usage Guidelines, allowing limited research use while prohibiting training or redistribution of outputs for commercial purposes.
\end{itemize}

All training and evaluation data were sourced from publicly available repositories or licensed datasets with appropriate authorization. No private, proprietary, or personally identifiable information (PII) was included. Our work is compliant with the EU General Data Protection Regulation (GDPR) and prevailing ethical norms.

\subsection*{Ethical Considerations and Oversight}

We recognize the dual-use nature of personality modeling. Such techniques may be misappropriated for impersonation or social engineering. As a precaution, we strongly discourage any deceptive or manipulative application of personality-simulating models and urge the broader community to develop and adopt transparent safety mechanisms. Public disclosure of model usage contexts and safeguards is essential to ensure responsible deployment.

The open-source release of all developed resources, including data processing pipelines, prompts, evaluation code, and bias analysis methodology, is licensed under the MIT License. Documentation will accompany the release to promote reproducibility and transparency.

Finally, we commit to ongoing ethical oversight, including periodic reevaluation of datasets, generation behaviors, and potential risks. Our efforts are aligned with environmental sustainability objectives, including the reduction of computing-related emissions in accordance with carbon neutrality goals.

\section{Limitations}
\begin{itemize}
    \item \textbf{Small, Coarsely Labeled Dataset:} The Essays Dataset \cite{pennebaker1999linguistic}, with 2,467 essays totaling 1.9 million words (average 770 words per essay), is larger and more narrative-rich than many datasets used in prior personality induction studies, such as Reddit posts \cite{gjurkovic2018reddit} or Twitter/X data \cite{golbeck2011predicting}, which often consist of thousands of short texts (e.g., 10,000–50,000 samples, typically under 280 characters). These smaller texts can lead to models learning superficial keyword patterns, as simple bag-of-words approaches like TF-IDF with SVMs already achieve comparable performance (60–80\% accuracy on individual traits) \cite{han2020knowledge, christian2021text}. The Essays Dataset’s longer, introspective narratives prioritize quality, capturing deeper personality signals, but its modest size may still limit linguistic diversity, and its binary Big Five labels may oversimplify trait gradients \cite{goldberg1993structure}. Moreover, high-quality, large-scale datasets with rich personality annotations are scarce, and generating synthetic data often oversimplifies the task by lacking the nuanced, context-dependent expressions found in human-generated text \cite{hamalainen2023evaluating, miotto2022gpt}. As mentioned earlier, if we mix smaller datasets consisting of tweets, Reddit posts, etc, into our work, it has a thread of polluting the results without achieving the original task of capturing personality, as short texts tend to be more "mood" driven.

    \item \textbf{Questionnaire Based Evaluation Bias:} While being grounded in psychology literature, Big Five questionnaires were intended for humans to self-report, and thus the low accuracy might reflect an inherent mismatch between introspection capabilities in humans vs. probabilistically generated tokens\cite{podsakoff2012sources, gow2005goldberg}.
    
    \item \textbf{No Human Evaluation:}
    Our study did not include cross-verification of the generated text by human experts, leaving the subjective quality and authenticity of personality expression in the outputs unassessed. Human evaluation is critical for determining whether the model’s generated text convincingly reflects targeted personality traits, such as linguistic markers of extraversion or conscientiousness, and passes an "eye check" for human-like quality. Research indicates that human evaluators often exhibit bias against AI-generated text when aware of its origin, rating it less favorably compared to human-authored content \cite{christian2021text}. Despite this potential bias, human validation would have provided valuable insights into whether our model’s outputs align with psychological expectations of personality expression. Given the negative results of our experiments, which showed limited success in inducing consistent personality traits, and the economic burden of recruiting expert evaluators, we refrained from conducting this resource-intensive task.

    \item \textbf{Threat of Overfitting}
    The small scale of the Essays Dataset and the multiplicative use of data in post-preferential fine-tuning increase the risk of overfitting, where the model may memorize patterns rather than generalize personality traits.

\end{itemize}

\section*{References}\label{sec:reference}
\bibliographystyle{lrec2026-natbib}
\bibliography{bibliography}

\FloatBarrier
\appendix

\section{Code-Grounded Reproducibility Details}
\label{app:code-grounded}

To facilitate reproducibility, this appendix reports the key implementation details extracted directly from the released codebase.\footnote{Available at \url{https://github.com/pkrajput/personality_induction}} We document the training hyperparameters, inference configuration, prompt templates, and dataset statistics used across all experiments.

\subsection{Training Configuration}
\label{app:training-config}

Table~\ref{tab:code-hparams} summarizes the default hyperparameters for both supervised fine-tuning (SFT) and preference-based training (DPO/ORPO) as specified in \texttt{train.py}. Notably, SFT employs standard LoRA on a smaller set of target modules, while DPO and ORPO use QLoRA with 4-bit quantization and a broader set of adapter targets to accommodate the additional memory overhead of pairwise training.

\begin{table}[H]
\centering
\small
\resizebox{\columnwidth}{!}{%
\begin{tabular}{lcc}
\toprule
\textbf{Parameter} & \textbf{SFT} & \textbf{DPO / ORPO} \\
\midrule
LoRA rank ($r$) & 8 & 16 \\
LoRA alpha ($\alpha$) & 32 & 32 \\
LoRA dropout & 0.1 & 0.05 \\
Target modules & \texttt{q\_proj, v\_proj} & \texttt{up/down/gate/k/q/v/o\_proj} \\
Learning rate & $1\times10^{-5}$ & $8\times10^{-6}$ \\
Optimizer & \texttt{adamw\_hf} & \texttt{adamw\_hf} \\
LR scheduler & TRL default & \texttt{linear} \\
Warmup steps & 500 & 500 \\
Batch size (train/eval) & 2 / 2 & 2 / 2 \\
Gradient accum.\ steps & 4 & 4 \\
Max sequence length & 4000 & 4000 \\
Epochs & 3 & 3 \\
Quantization & FP16 & 4-bit \textit{nf4} \\
\midrule
ORPO $\beta$ & -- & 0.1 \\
\bottomrule
\end{tabular}%
}
\caption{\centering Training hyperparameter defaults extracted from \texttt{train.py}. SFT uses standard LoRA while DPO/ORPO use QLoRA with broader adapter coverage.}
\label{tab:code-hparams}
\end{table}

\FloatBarrier
\subsection{Inference Configuration}
\label{app:inference-config}

Table~\ref{tab:code-inference} lists the inference-time defaults from \texttt{inference.py}. All evaluations use greedy decoding (temperature~$=0$) to minimize stochastic variation between runs. The trait threshold of 3.0 on a 1--5 scale determines the binary classification boundary for each OCEAN dimension.

\begin{table}[H]
\centering
\small
\begin{tabular}{p{0.55\columnwidth}l}
\toprule
\textbf{Flag / Setting} & \textbf{Default} \\
\midrule
\texttt{-{}-prompt\_set} & \texttt{S1} \\
\texttt{-{}-num\_profiles} & 32 \\
\texttt{-{}-question\_split} & \texttt{test} \\
\texttt{-{}-with\_essay} & off (flag) \\
Questionnaire generation & greedy \\
Questionnaire \texttt{max\_new\_tokens} & 200 \\
Essay \texttt{max\_new\_tokens} & 500 \\
Trait threshold & 3.0 \\
\bottomrule
\end{tabular}
\caption{\centering Inference defaults from \texttt{inference.py}. Greedy decoding is used throughout to ensure deterministic outputs.}
\label{tab:code-inference}
\end{table}

\FloatBarrier
\subsection{Effect of Temperature on Per-Trait Accuracy}
\label{app:temperature}

\begin{figure}[H]
\centering
\includegraphics[width=\columnwidth]{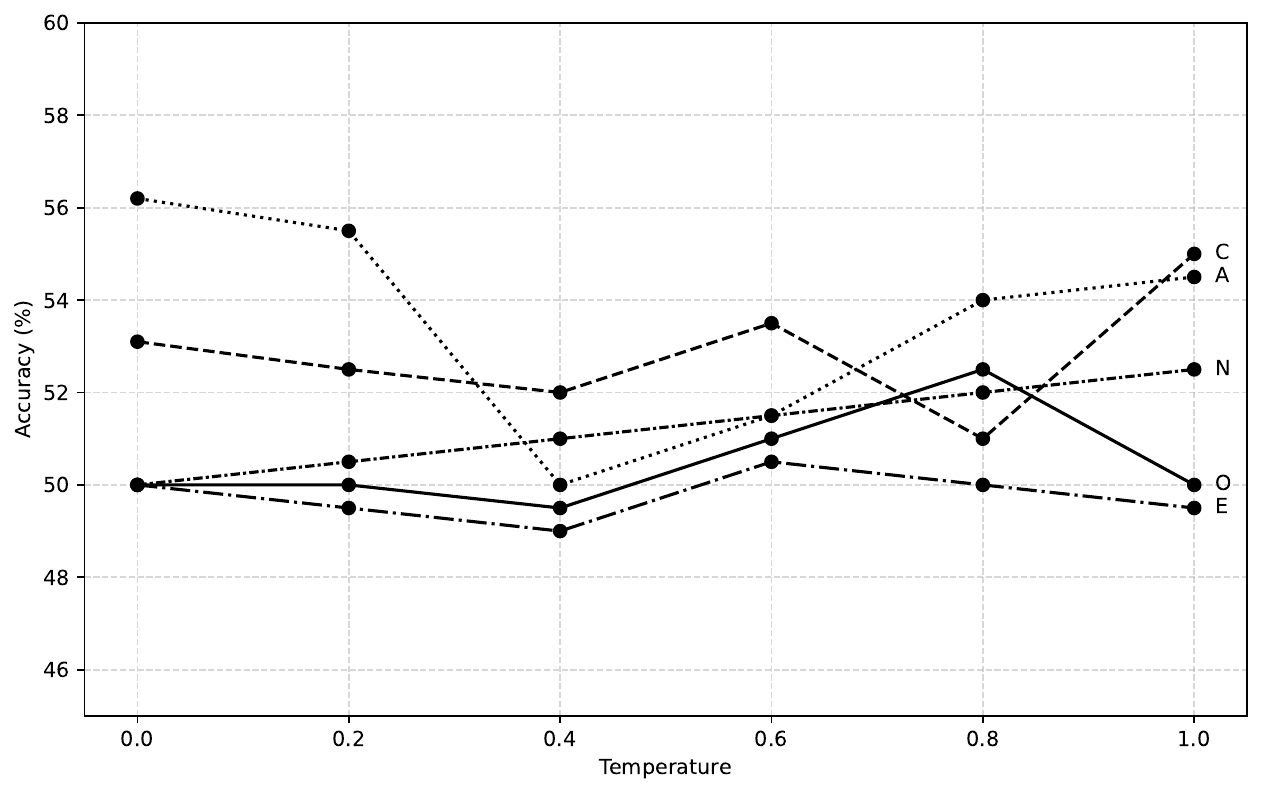}
\caption{Per-trait OCEAN accuracy for GPT-3.5-turbo-0125 (SFT) across decoding temperatures. Accuracy varies by $\leq 6\%$, confirming that temperature has a minor effect on trait-level scores.}
\label{fig:temperature-ablation}
\end{figure}

Figure~\ref{fig:temperature-ablation} shows per-trait accuracy for GPT-3.5-turbo-0125 (SFT) as a function of decoding temperature. Accuracy remains largely stable across temperature values, with deviations of $\leq 6\%$ from greedy decoding, supporting our choice of temperature~$=0$ for all main experiments (cf.\ Section~4.3).

\FloatBarrier
\subsection{Prompt Templates}
\label{app:prompts}

The codebase uses format-specific prefixes to parse model responses reliably. For the numeric prompt set (S1), the response prefix is: \texttt{My score for the statement is:}, while for the string and alphabetical sets (S2/S3), the prefix is: \texttt{My answer to the statement is:}. These prefixes enable straightforward extraction of structured responses from free-form model output.

For essay generation during evaluation, the following template is used:

\begin{tcolorbox}[colback=gray!5,colframe=gray!15,boxsep=1pt,left=4pt,right=4pt,top=3pt,bottom=3pt]
\small
\texttt{Write a free-hand essay about your feelings in the moment as if you are a person who is \{trait\_desc\}.}
\end{tcolorbox}

\noindent Here, \texttt{\{trait\_desc\}} is dynamically populated with the target personality description (e.g., ``positive in openness, negative in conscientiousness, \ldots'') corresponding to one of the 32 possible Big Five binary combinations.

\FloatBarrier
\subsection{Dataset Statistics}
\label{app:data-stats}

Table~\ref{tab:data-counts} reports the sample counts for every data artifact in the repository, grouped by training paradigm. The base SFT split contains 1,652 training samples derived from the Essays Dataset after OpenAI moderation filtering ($\approx$300 samples removed), following an 80/10/10 train/validation/test ratio. The \emph{with questionnaires} variants double these counts because each essay appears twice: once with deterministic ideal questionnaire responses and once with randomized responses whose trait-level means match the target personality. The DPO/ORPO preference datasets are approximately three times the base training size, reflecting the triplet sampling strategy (3 rejected alternatives per chosen essay) described in Section~4.2.3. Preference data is constructed for both the base essays and the questionnaire-augmented variants.

\begin{table}[H]
\centering
\small
\begin{tabular}{llr}
\toprule
\textbf{Training Paradigm} & \textbf{Split} & \textbf{Samples} \\
\midrule
\multirow{3}{*}{SFT (essays only)} & Train & 1,652 \\
 & Validation & 207 \\
 & Test & 207 \\
\midrule
\multirow{3}{*}{\shortstack[l]{SFT (essays +\\questionnaires)}} & Train & 3,304 \\
 & Validation & 414 \\
 & Test & 414 \\
\midrule
\multirow{2}{*}{\shortstack[l]{DPO / ORPO\\(preference pairs)}} & Train & 5,577 \\
 & Test & 621 \\
\midrule
\multirow{2}{*}{\shortstack[l]{DPO / ORPO\\(with questionnaires)}} & Train & 11,154 \\
 & Test & 1,242 \\
\bottomrule
\end{tabular}
\caption{Dataset statistics grouped by training paradigm. The base SFT split uses an 80/10/10 ratio from the filtered Essays Dataset. The questionnaire-augmented SFT variant doubles each split (ideal + randomized responses). DPO/ORPO triples the datasets via 3 rejected alternatives per chosen essay, including both base and questionnaire-augmented variants.}
\label{tab:data-counts}
\end{table}

\FloatBarrier
\subsection{Training Input Structure by Variant}
\label{app:input-structure}

Table~\ref{tab:input-structure} shows how the training input is assembled for each experimental variant. SFT variants use a chat-style format (system/user/assistant), while DPO/ORPO uses a flat prompt/chosen/rejected structure consumed directly by the TRL preference trainers.

\begin{table*}[h!]
\centering
\small
\renewcommand{\arraystretch}{1.15}
\begin{tabular}{p{2.4cm}p{1.4cm}p{10.8cm}}
\toprule
\textbf{Variant} & \textbf{Role} & \textbf{Content} \\
\midrule
\multirow{3}{2.4cm}{SFT\\(essays only)}
 & System & \texttt{You are writing an essay that mimics the personality of real humans. You will be given a binary requirement for their Big Five traits.} \\
 & User & \texttt{Write an essay as a person \{O\} in openness, \{C\} in conscientiousness, \{E\} in extroversion, \{A\} in agreeableness, and \{N\} in neuroticism.} \\
 & Assistant & \textit{$\langle$human-authored essay$\rangle$} \\
\midrule
\multirow{3}{2.4cm}{SFT\\(essays +\\questionnaires)}
 & System & \texttt{You are writing an essay that mimics the personality of real humans. You will be given a binary requirement for their Big Five traits along with their questionnaire responses.} \\
 & User & \texttt{Write an essay as a person \{O\} in openness, \ldots, and \{N\} in neuroticism.}\newline\newline\texttt{The following are the person's responses to personality questionnaire items:}\newline\texttt{[Agreeableness Questionnaire Responses]}\newline\texttt{~~Q: Accept people as they are. -> Score: 1}\newline\texttt{~~Q: Am annoyed by others' mistakes. -> Score: 5}\newline\texttt{~~\ldots}\newline\texttt{[Extraversion Questionnaire Responses]}\newline\texttt{~~\ldots}\newline\textit{(repeated for all five OCEAN dimensions using the train-split items)} \\
 & Assistant & \textit{$\langle$human-authored essay$\rangle$} \\
\midrule
\multirow{3}{2.4cm}{DPO / ORPO\\(preference\\pairs)}
 & Prompt & \texttt{Write a free-hand essay about your feelings in the moment as if you are a person who is \{E\} in extraversion, \{N\} in neuroticism, \{A\} in agreeableness, \{C\} in conscientiousness, and \{O\} in openness.} \\
 & Chosen & \textit{$\langle$essay matching the target OCEAN profile$\rangle$} \\
 & Rejected & \textit{$\langle$essay with a different, non-matching OCEAN profile$\rangle$} \\
\midrule
\multirow{3}{2.4cm}{DPO / ORPO\\(with\\questionnaires)}
 & Prompt & \texttt{Write a free-hand essay about your feelings in the moment as if you are a person who is \{E\} in extraversion, \{N\} in neuroticism, \{A\} in agreeableness, \{C\} in conscientiousness, and \{O\} in openness.}\newline\newline\texttt{The following are the person's responses to personality questionnaire items:}\newline\textit{(same per-trait Q\&A format as SFT + questionnaires)} \\
 & Chosen & \textit{$\langle$essay matching the target OCEAN profile$\rangle$} \\
 & Rejected & \textit{$\langle$essay with a different, non-matching OCEAN profile$\rangle$} \\
\bottomrule
\end{tabular}
\caption{\centering Training input structure for each experimental variant. \texttt{\{O\}}, \texttt{\{C\}}, \texttt{\{E\}}, \texttt{\{A\}}, \texttt{\{N\}} denote the binary trait descriptors (\texttt{positive}/\texttt{negative}). SFT uses chat-style messages consumed by the tokenizer's chat template; DPO/ORPO uses flat prompt/chosen/rejected fields processed by TRL's preference trainers. The questionnaire-augmented variants append per-trait Q\&A items from the train split of the IPIP-NEO inventory to the prompt.}
\label{tab:input-structure}
\end{table*}

\end{document}